# Pragmatic classification of movement primitives for stroke rehabilitation

**Avinash Parnandi**[*]
NYU School of Medicine
New York, United States
avinash.parnandi@nyulangone.org

**Jasim Uddin**
Columbia University Medical Center
New York, United States
ju2189@cumc.columbia.edu

**Dawn M. Nilsen**
Columbia University Medical Center
New York, United States
dmn12@cumc.columbia.edu

**Heidi Schambra**
NYU School of Medicine
New York, United States
heidi.schambra@nyulangone.org

*Abstract*

**Background**. Rehabilitation training is the primary clinical intervention to improve motor recovery after stroke, but a tool to measure functional training dose in the upper extremities (UE) does not currently exist. To bridge this gap, we previously developed an approach to classify functional movement primitives using wearable sensors and a machine learning (ML) algorithm. We found that this sensor-ML approach had encouraging classification performance but had computational and practical limitations, such as ML training time and sensor cost and electromagnetic drift. In this study, we sought to refine this approach to facilitate real-world implementation. We determined the ML algorithm, sensor configurations, and data requirements needed to maximize computational and practical performance.

**Methods.** Motion data had be previously collected from six stroke patients wearing 11 inertial measurement units (IMUs) as they moved objects on a horizontal target array. To identify optimal ML performance, we evaluated four off-the-shelf algorithms that are commonly used in activity recognition (linear discriminant analysis (LDA), naïve Bayes classifier, support vector machine, and k-nearest neighbors). We compared their classification accuracy, computational complexity, and tuning requirements. To identify optimal sensor configuration, we progressively sampled fewer sensors and compared classification accuracy on reduced datasets. To identify optimal data requirements, we compared classification accuracy using data from IMUs versus accelerometers.

**Results.** We found that LDA had the highest classification accuracy (positive predictive value (PPV) 92%) of the ML algorithms tested. It also was the most pragmatic, with low training (26 s) and testing times (0.04 ms) and modest tuning requirements. We found that seven sensors (paretic hand, forearm, arm, sternum, pelvis, and scapula) resulted in the best accuracy (PPV 92%). Using this array, accelerometry data produced a lower accuracy (PPV 84%) than IMU data.

**Conclusions.** Here, we refined strategies to accurately and pragmatically quantify functional movement primitives in stroke patients. From the computational perspective, LDA represented the best balance of performance and practicality. From the sensor perspective, seven IMUs on the paretic limb and trunk enabled the best classification accuracy. We propose that this optimized ML-sensor approach could be a means to quantify training dose after stroke.

[*]Corresponding author: Avinash Parnandi (avinash.parnandi@nyulangone.org)

## *1. Introduction*

Over six million stroke survivors in the US have upper extremity (UE) motor impairment, resulting in a loss of independence that costs over $27 billion annually [1-3]. To promote UE recovery in the weeks-months after stroke, patients undergo rehabilitation, which commonly focuses on functional object use in the context of activities of daily living (ADLs).

Studies in lesioned animals have found that if functional movements are trained early enough and are given in sufficiently high quantity, robust motor recovery can be achieved [4-6]. In human rehabilitation, optimal training doses are unknown, in part because a pragmatic measurement tool to identify *what* and *how much* is being trained during rehabilitation does not currently exist.

A first step in dosing rehabilitation is to identify a standard unit of measure. We decompose functional activities into movement primitives—discrete, object-oriented motions with a single goal. We focus on movement primitives because they: (1) are non-divisible and are largely invariant across individuals [7], (2) may be represented at the cortical level [8], and (3) provide a finer-grained capture of performance in stroke patients who may be unable to accomplish a full activity. Like phonemes, movement primitives can be strung together in various combinations to make a functional movement [9] (analogous to a word), which in turn are strung together to make a functional activity (analogous to a sentence) [7]. For example, a series of *reach-transport-reach* primitives could constitute a functional movement for opening a bottle cap, within the activity of drinking.

Previous attempts to quantify rehabilitation dose have been limited by imprecision. Most neurorehabilitation studies use time scheduled for therapy as a proxy [10, 11], which fails to capture both training content and quantity [12]—paramount for translating findings to clinical practice. Other attempts to quantify dose have been limited by impracticality. Observation-based approaches, such as manual counting or computer vision, require an unobstructed line of sight, multiple viewing angles, and/or laborious review of video, making them unrealistic for rehabilitation environments [13].

Wearable sensors, such as inertial measurement units (IMUs) and accelerometers, provide rich and continuous kinematic data and allow seamless motion capture—important for clinical applications. We recently used IMUs to quantify movement primitives in stroke subjects performing a structured tabletop activity. We applied a machine learning (ML) approach (hidden Markov model-logistic regression) to recognize movement primitives embedded in this task, finding an overall classification accuracy of 79%. [14] However, this sensor-ML approach had variable classification performance among the primitives (62-87% accuracy). It also did not address implementation challenges, such as the level of domain knowledge required, the computational costs, or the expense and electromagnetic intolerance of IMUs.

In the present study, we addressed the limitations of the sensor-ML approach by optimizing movement capture and analysis capabilities. We compared several ML algorithms, sensor configurations, and data requirements to maximize the computational and practical performance of our approach. An approach with high classification performance, low computational complexity, and low practical restrictions would bring rehabilitation dose quantitation closer to reality.

## 2. Methods

The current study leverages data collected in previous work [14]. We briefly describe the experimental setup here. Six mild-to-moderately impaired stroke patients (Table 1) moved a toilet paper roll and aluminum can over a horizontal array of targets (Fig. 1).

| N | 6 |
|---|---|
| Age (years) | 61.7 (46.5 -71.0) |
| Gender (Female/Male) | 2F/4M |
| Race (Asian/Black/White) | 1A/2B/3W |
| Dominant arm (Right/Left) | 5R/1L |
| Paretic side (Right/Left) | 6R |
| Impairment (Fugl-Meyer score) | 52.8 (45 - 62) |
| Time since stroke (years) | 12.0 (2.0 - 31.1) |

**Table 1. Demographic and clinical characteristics of patients.** Shown are number of participants, mean age (range), gender, race, hand dominance, paretic side, mean Fugl-Meyer assessment score at first assessment (range; maximum 66), and time since stroke (range)**.** Inclusion criteria were age ≥18 years; premorbid right-hand dominance; unilateral motor stroke; contralateral arm weakness with Medical Research Council score <5/5 in a major muscle group. Exclusion criteria were traumatic brain injury; musculoskeletal, medical, or non-stroke neurological condition interfering with assessment of motor function; contracture at shoulder, elbow, or wrist; moderate dysmetria or truncal ataxia; visuospatial neglect; apraxia; global inattention; blindness.

Subjects performed 5 trials moving the object between a center target and eight radially arrayed targets (20 cm away). The task generates the following movement primitives: *reach* (movement from *idle* to grasping object); *transport* (movement conveying object); *reposition* (releasing object and returning UE to *idle*); and *idle* (minimal movement in UE). To record movement data, subjects wore 11 IMUs (XSens Technology) placed on head, sternum, pelvis, and bilateral hands, forearms, arms, and scapulae. IMUs capture linear acceleration and angular velocity, and the XSens software computes quaternions, at 240 Hz. To segment and label the motion data as constituent primitives, we synchronously recorded movement (30 Hz) with a single video camera.

Trained coders used the video recording to label the beginning and end of each movement primitive, which also labeled the corresponding IMU data. This step enabled us to train ML algorithms on motion data and test their classification performance against a ground-truth label. Data were pre-processed by extracting statistical features prior to feeding it to the machine algorithms. We extracted statistical features including mean, standard deviation, minimum, maximum, entropy, skewness, energy, and root mean square to characterize the IMU data. These statistical descriptors have been shown to capture human movement [15-17]. Following prior work, we selected a window size of 0.25s sliding by 0.1s [14]. Data were z-score normalized before computing the features. The dataset consisted of 810 *reaches*, 708 *transports*, 781 *repositions*, and 582 *idles*.

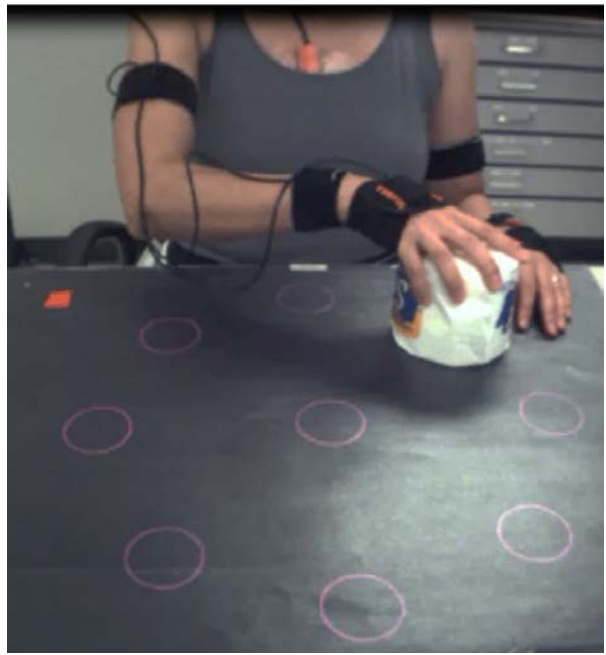

**Figure 1. Tabletop task set-up.** Healthy individual wearing the sensors and transporting the object from center to a target in the functional task.

*3. Computational details*

**3.1 Machine learning (ML) methods for classification**.

In the present study, we sought to identify an ML algorithm that performs well, i.e. has a high classification accuracy, but that also is practical, i.e. has a low computational overhead and minimal tuning requirements.

Supervised ML algorithms work in two phases: training and testing. During training, ML algorithms learn the relationship between a pattern of data characteristics (here, the statistical features) and its class (here, its movement primitive label). During testing, the trained ML algorithm uses the pattern of data characteristics to identify a fresh data sample as one of the primitives. This identification is checked against the human label, thus reading out classification accuracy.

We considered generative and discriminative algorithms. Generative methods model the underlying distribution of data for each class, seeking to identify data characteristics that enable matching of new data samples to a given class. Generative algorithms include linear discriminant analysis, naïve Bayes classifier, and hidden Markov model. In contrast, discriminative methods model the boundaries between classes and not the data themselves. They seek to identify the plane separating the classes so that, based on location relative to the plane, a new data sample is assigned to the appropriate class. Discriminative algorithms include support vector machines, k-nearest neighbors, and logistic regression.

We selected four algorithms that have been found to provide high classification performance in human activity recognition: linear discriminant analysis (LDA) [16], Naïve Bayes classifier (NBC) [15], support vector machine (SVM) [18], and k-nearest neighbors (KNN) [17]. We used "off the shelf" versions of these algorithms without any special permutations; in other words, the algorithms are widely available in most machine learning libraries [19, 20]. Their computational characteristics are summarized briefly below.

**Linear discriminant analysis.** LDA projects training data to a lower dimension that maximizes the separation between classes [21]. During algorithm training, projection vectors are computed that maximize the ratio of between-class scatter and within-class scatter. This can be transformed into an optimization problem as follows:

$$\min_{w} -\frac{1}{2} w^T S_B w : w^T S_w w = 1$$

where $S_B$ is the between-class scatter matrix and $S_w$ the within-class scatter matrix. This results in the projection vectors $w$ (and its transpose $w^T$) to project the data into a lower dimensional space. During algorithm testing, a new sample is projected into this lower dimensional space and is assigned to the class with the lowest Mahalanobis distance.

**Naïve Bayes classifier.** NBC uses Bayes' rule and prior information to classify a new sample $x_i$ [22]. During algorithm training, NBC estimates the prior probability of each class in the dataset ($C_k$) and the distribution (mean and standard deviation) of features in a class $p(x_i|C_k)$. During algorithm testing, NBC computes the posterior probability—the change in prior belief given new information—of a new sample $x_i$ as follows:

$$y = \underset{k \epsilon \{1,...,k\}}{\mathrm{argmax}}\, p(C_k) \prod_{i=1}^{n} p(x_i \mid C_k)$$

In other words, given a new sample, NBC computes membership probabilities for each class—the probability that the new sample belongs to a particular class. The class with the highest probability $y$ is taken as most likely, and the sample is assigned to that class.

**Support vector machine.** SVM is based on the idea that increasing the dimensionality of data makes their classification easier. SVM discriminates between data classes by finding a hyperplane that separates them [23]. During algorithm training, training data are projected to a high dimensional space using a non-linear function. A hyperplane with maximum distance from the training data belonging to the two classes is computed as follows:

$$\min ||w^T w|| \;:\; y_i(w^T x_i + b) \geq 1 \text{ for } i = 1 \text{ to } n$$

where $x$ are the training samples, $w$ and $b$ are the weight and bias for the hyperplane, and $y$ is the class label. During algorithm testing, a new sample is projected in the high dimensional space and classified based on location relative to the hyperplane. For example, a new sample will be assigned to one class if it is above the hyperplane and to another class if it is below. SVM by default is restricted to binary classification. We trained four independent SVMs in a one-versus-all design and used these for identifying the primitives [24].

**K-nearest neighbors.** In contrast to the other algorithms, KNN does not require a training phase [25]. Rather, KNN relies on the assumption that samples from the same class will share similar data characteristics. During algorithm testing, the distances between a test sample's features and those of a predetermined number of the closest data samples ('k') are computed using a Euclidean distance metric. The test sample is assigned to the class with the majority of closest distances. Based on prior work in activity recognition, we chose k = 5 in our implementation [17].

**3.2 Algorithm performance metrics.**

**Classification performance of algorithms.** We first evaluated how well the algorithms could classify primitives in the dataset. We used 60% of the data to train the algorithm and 40% to test it, repeating the process 10 times. Data were randomly selected for each primitive proportional to its prevalence in the complete dataset (i.e., stratified proportional sampling). This ensured that each data subset adequately represented the entire sample population.

In the algorithm testing phase, we estimated classification accuracy by comparing algorithm-chosen labels against the ground truth of human labels. We used positive predictive value (PPV) as the performance metric. Comparing algorithm labels against human labels, primitives were classified as true positive (*TP*, labels agreed) and false positive (*FP*, labels disagreed), generating the PPV (*TP/(TP+FP)*) of the algorithm. PPV reflects how often a primitive was actually performed when the algorithm labeled it as such; in other words, PPV is how often a primitive was correctly classified. We generated primitive-level PPVs in a one-versus-all analysis (e.g., *reach* vs. *transport + reposition + idle* combined). We also generated an overall PPV by combining data for all primitives and tallying all true and false positives. We prefer PPV because it takes into account the prevalence of the primitive in the dataset [26].

We also used receiver operating characteristic (ROC) curves to assess the classification performance of the algorithms. ROC curves are generated using a one-versus-all analysis and drawn with true positive rate (TPR) as the x-axis and false positive rate (FPR) as the y-axis. TPR (or sensitivity) represents the number of correct classifications given the primitive was actually made. FPR (or 1-specificity) refers to the number of incorrect classifications given the movement primitive was not performed. Therefore, ROC curves depict the relative tradeoff between sensitivity and specificity and identify the optimal operating point of an algorithm, indicating the best tradeoff between sensitivity and specificity. The operating point is useful in selecting a classifier with desired characteristics; for example, one that favors high true positives and low false positives will be a good candidate for primitive identification. Perfect classification would lead to a ROC curve that passes through the upper left corner, with an area under the ROC curve (AUC) equal to 1, and an operating point of 100% sensitivity and 100% specificity [27].

**Practical performance of algorithms.** We next considered the pragmatic implementation of each algorithm by assessing their computational complexity. First, we estimated the time required to train and test the algorithms on datasets of different sizes, using data randomly selected from our dataset (20-100% of the dataset in 10% increments). For each dataset size, we measured the time required to train the algorithm, and the time required for a fully trained algorithm to classify a primitive. For each dataset size, the algorithms were trained *de novo* to avoid overfitting and to provide unbiased estimates. A fast training time enables the rapid appraisal of classification performance, allowing an investigator to select the appropriate algorithm or to iteratively optimize its parameters. A fast testing time favors implementation in a clinical setting by generating real-time classification of primitives.

We additionally assessed the real-world ramifications of algorithm training and testing time for a dataset collected in typical paradigm (NS104207). We generated a simulated dataset of 300,000 primitives with same proportion, mean, and variance as our original dataset. We assumed that a computer executes one billion computations per second [28]. To measure simulated training times, we estimated times to process 25-100% of the simulated dataset in increments of 25%. To measure simulated testing times, we estimated the time required for a fully trained algorithm to classify a primitive.

Second, we assessed the algorithm's need for tuning, which is the informed adjustment of algorithm parameters in order to maximize classification performance. While our algorithms were applied "off the shelf," each allows for parameter tuning. We operationalized this tuning requirement as the number of parameters that can be adjusted. We also qualitatively classified the level of domain knowledge typically required to implement and tune the algorithms, where "low" indicates a basic knowledge of statistics, "medium" indicates undergraduate-level knowledge of machine learning, and "high" indicates graduate-level knowledge of machine learning. Of note, this scale is based on typical US educational programs, but given sufficient didactics, a motivated undergraduate could achieve a "high" level of knowledge.

**Optimal data characteristics**. We then focused on the hardware side of our approach, seeking to identify the best balance between ease of data capture and high classification performance. The IMU system generates 3D linear accelerations, 3D angular velocities, and 4D quaternions, resulting in 10 data dimensions per sensor. However, IMUs have practical limitations that include electromagnetic drift and cost. Magnetic environments lead to potentially inaccurate IMU-derived motion estimates and a need for frequent recalibration. On the other hand, 3D accelerometers generate only linear acceleration data, resulting in 3 data dimensions. However, accelerometers are inexpensive and are largely unaffected by magnetic environments. Although simplified motion capture would favor clinical implementation, sparser data may reduce classification performance.

In this analysis, we identified the minimal number, configuration, and type of sensor that could still maintain a high classification performance. For this analysis, we subsampled data from the full IMU data stream, thus ensuring identical sensor locations and analyzed movements for comparison. We used LDA to generate classification metrics because it performed best in the analyses above.

We first evaluated IMU number and configurations, identifying the minimal number of IMUs and their location on the body that could support modest classification accuracy. We selected IMU number and configurations using domain knowledge and exhaustive search. With domain knowledge, we used clinical judgment to progressively remove IMUs; for example, we expected that in the unimanual task, IMUs on the non-active arm could be removed without a significant loss in classification performance. By contrast, with exhaustive search, all possible IMU configurations were systematically evaluated [29]. Given that exhaustive search assesses all IMU configurations, it provides an unbiased validation of results achieved using domain knowledge.

Second, we evaluated which type of data optimized algorithm performance. We compared classification accuracies using IMU data versus accelerometry-only data. This analysis allowed us to determine whether accelerometry data, with its reduced dimensionality, could be used in lieu of IMU data to achieve a sufficiently high classification accuracy.

## *4. Results*

**4.1 Classification performance of algorithms.**

We first determined the classification performance of multiple ML algorithms using PPVs (Table 2), indicating how often an algorithm correctly identified a primitive. LDA and SVM had high classification performance for all primitives (overall PPV 92.5% and 92%, respectively). KNN had intermediate performance (PPV 87.5%) and NBC had the lowest performance (PPV 80.2%), particularly for *reaches* (PPV 77%) and *transports* (PPV 71%).

| Algorithm | PPVs for functional movement primitives | | | | Overall PPV |
|---|---|---|---|---|---|
| | Reach | Transport | Reposition | Idle | |
| LDA | 93% | 91% | 93% | 92% | 92.5% |
| NBC | 77% | 71% | 83% | 85% | 80.2% |
| SVM | 92% | 90% | 92% | 93% | 92% |
| KNN | 86% | 87% | 85% | 89% | 87.5% |

**Table 2. Classification performance of machine learning algorithms for movement primitives.** Positive predictive value (PPV), which reflects how often a primitive was actually made when the algorithm identified it as such, was calculated for the primitives of *reach*, *transport*, *reposition*, and *idle*. Primitive-level PPVs were computed in one-versus-all analysis (e.g., *reach* vs. *transport* + *reposition* + *idle* combined). The overall PPV was assessed by combining data for all primitives and tallying all true and false positives. Overall classification performance was highest for linear discriminant analysis (LDA) and support vector machine (SVM), moderately high for k-nearest neighbors (KNN), and lowest for Naïve Bayes classifier (NBC).

To further characterize classification performance, we generated ROC curves for each primitive (Fig. 2). All algorithms detected *idle* with high accuracy (AUC > 0.87). For the other primitives, LDA and SVM had AUCs 0.95-0.99, indicating very high classification accuracy. KNN also had high classification accuracy for *reach* (AUC 0.94) and *transport* (AUC 0.90) and intermediate classification accuracy for *reposition* (AUC 0.87). In contrast, NBC had the lowest classification accuracy on the remaining primitives (AUC 0.80-0.85). We also identified the optimal operating point, indicating the best tradeoff between sensitivity and specificity, for each algorithm (Fig. 2). At their respective optimal operating points, LDA and SVM achieved high sensitivities (0.83-0.95) and specificities (0.83-0.95) for all primitives. KNN achieved a high sensitivity (0.91) and specificity (0.86) for *transport*, but had moderate sensitivities (0.80-0.88) and specificities (0.79-0.86) for other primitives. NBC had the lowest sensitivities (0.74-0.81) and specificities (0.74-0.79) for all primitives. In sum, these findings indicate that LDA and SVM have the highest classification performance of the algorithms tested.

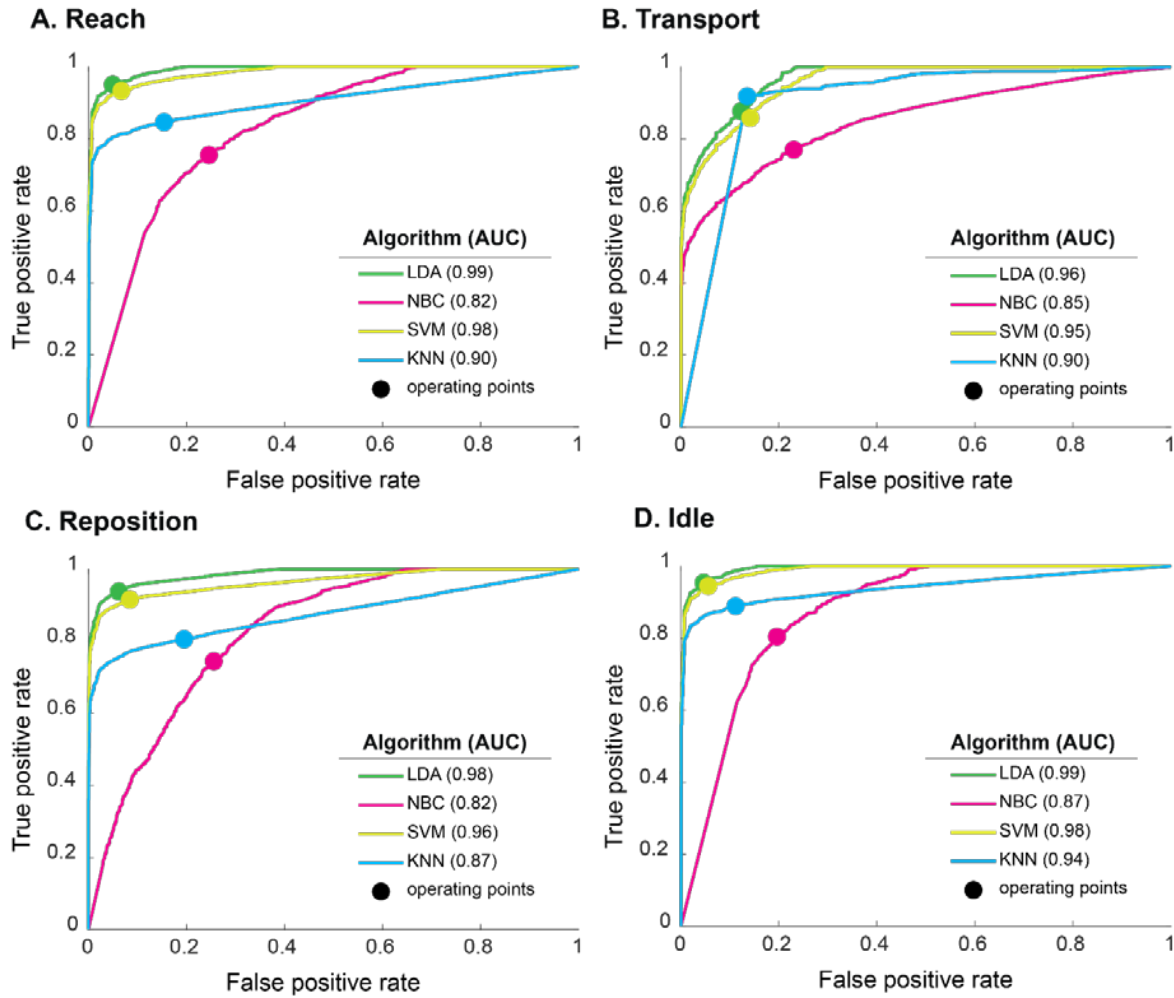

**Figure 2. Performance characteristics of machine learning algorithms for (A) Reach, (B) Transport, (C) Reposition, and (D) Idle.** Receiver operating characteristic (ROC) curves show the trade-off between true positive rate (or sensitivity) and false positive rate (1-specificity). Curves closer to the top-left corner indicate a better classification performance. The optimal operating point for each algorithm (solid circles), reflect the best tradeoff between sensitivity and specificity for an algorithm. The area under the curve (AUC), a measure of classification accuracy, is shown in parenthesis for each algorithm. AUC=1 represents perfect classification. LDA had the highest AUCs followed closely by SVM, indicating high classification performances. NBC had consistently the lowest AUCs, indicating the weakest classification performance.

**4.2 Practical performance of algorithms.** We next evaluated pragmatic aspects of algorithm implementation, to gauge real-world applicability. We calculated the time required to train and test the algorithm on increasing quantities of data (Fig. 3) from our dataset of 2880 primitives. SVM required the longest to train, on the order of minutes (5.6 min), with training times growing quadratically with increasing data quantity. Training times for NBC and LDA were on the order of seconds (12 s and 26 s, respectively), with training times growing linearly with increasing data quantity. As an inherent property of the model, KNN required no time to train. For the dataset of 2880 primitives, KNN required the longest to classify a new primitive (1.5 ms), with testing times growing linearly with increasing dataset size. In contrast, LDA, NBC, and SVM required constant time (approximately 0.04 ms) for testing.

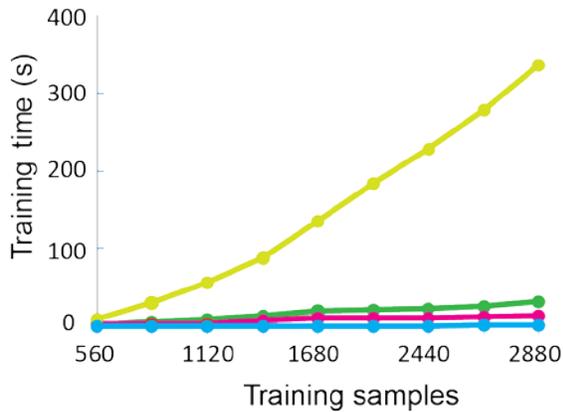
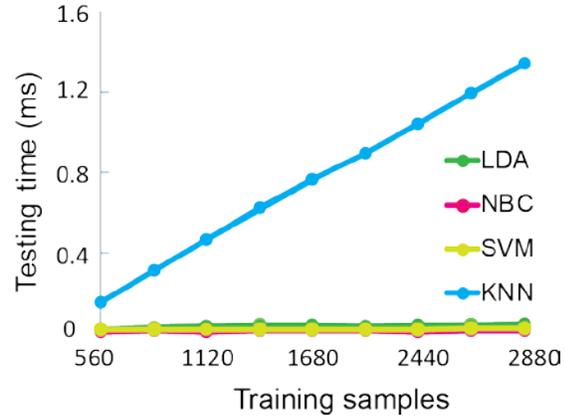

**Figure 3: Algorithm (A) training times and (B) testing times on sample dataset**. The dataset is comprised of 2880 primitives. We computed times to train and test each algorithm on 20-100% of the dataset in increments of 10%. To avoid overfitting and compute an unbiased estimate of training and testing times, ML algorithms were trained and tested *de novo* with each incremental increase. For training with the complete sample dataset, SVM required the most time (336 s) while the other algorithms finished training rapidly (<30 s). For testing, KNN required the most time (1.5 ms), while the other algorithms finished testing rapidly (~0.04 ms).

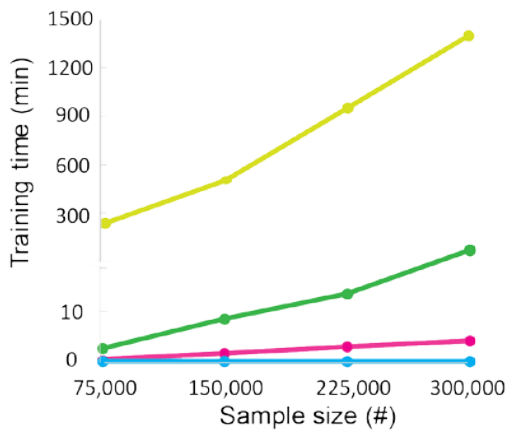
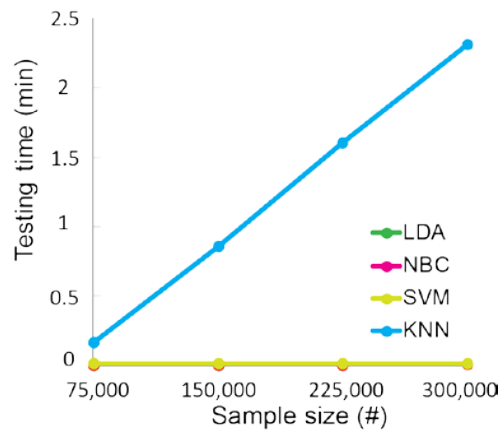

**Figure 4. Algorithm (A) training times and (B) testing times on real world-sized dataset**. The dataset is comprised of 300,000 simulated primitives. We evaluated training and testing times for quartile increases in dataset size. To avoid overfitting and compute unbiased estimates, the algorithms were trained and tested *de novo* at each quartile. For training with the entire dataset, SVM required the most time (1380 min) while the other algorithms required less time (LDA: 13 min; NBC: 2.5 min). Please note break in the y-axis to highlight the difference in the algorithm training times. For testing, KNN required the most time (2.3 min) while rest of the algorithms required much less time (~0.09 ms), which stayed the same with increased sample size.

We further assessed the ramifications of algorithm training and testing times in a real world-sized dataset, using 300,000 simulated primitives (Fig. 4). We found that training times became prohibitively long for SVM (up to 23 h) but were manageable for the other algorithms (up to 13 min). We found that the testing time for classifying a new movement primitive was relatively high for KNN (up to 2.3 min), whereas LDA, NBC, and SVM required a nominal and constant testing time (~0.09 ms). In sum, these findings indicate that LDA and NBC have the highest practical performance of the algorithms tested.

**4.3 Practical implementation of the algorithms.** Tuning requirements, which imply the complexity of algorithm implementation, are listed in Table 3. NBC has the lowest number of parameters (1) and requires the least amount of domain knowledge in machine learning to implement it. Although KNN has a moderate number of parameters (5), their optimization is reasonably intuitive and requires little domain knowledge. LDA has fewer parameters (3), but they require a higher level of domain knowledge. SVM has many parameters (9) and requires extensive domain knowledge to build an accurate and efficient model. In sum, these findings indicate that NBC and KNN are the simplest to implement, though LDA has only a few tuning parameters.

| Algorithm | # tuning parameters | Tuning parameters | Level of domain knowledge |
|---|---|---|---|
| LDA | 3 | Prior probability, regularization term, optimizer | Medium |
| NBC | 1 | selection of prior distribution | Low |
| SVM | 9 | Kernel function, kernel parameters (scale, offset), regularization term, # of iterations, Nu, prior probability, convergence parameter, optimizer | High |
| KNN | 5 | # of neighbors (K), distance metric, search algorithm, tie breaker, weighing criterion | Low |

**Table 3. Complexity of algorithm implementation.** Algorithm parameter tuning is necessary to achieve optimal classification performance. Shown are algorithm tuning characteristics, as indicated by number and specifics of the tuning parameters. Also shown is a graded estimate of the level of domain knowledge required to tune these parameters. NBC is considered the simplest to tune while SVM is the most difficult. LDA has a handful of parameters that require medium domain knowledge to negotiate. KNN has a moderate number of parameters that are intuitive to tune and require little domain knowledge. Level of domain knowledge: low, basic knowledge of statistics; medium, undergraduate-level knowledge of ML; high, graduate-level knowledge of ML.

**4.4 Optimal data characteristics.**

**Identifying optimal IMU configuration.** To evaluate the contribution of IMU number to primitive classification, we used domain knowledge and exhaustive search to progressively reduce IMU number and location. LDA was trained and tested on the progressively diminished dataset to read out effects on classification performance. With domain knowledge, we sequentially removed the scapula, arm, forearm, and hand IMUs from the non-active side (leaving seven IMUs on the head, sternum, pelvis, and UE of the active side). This improved classification performance from PPV 88% to 92.5% (Fig. 5). Next, we removed IMUs on the trunk and then head, given they generated less motion data than the four IMUs on the active UE. This reduced performance to PPV 81%. Finally, we progressively removed the scapula, arm, and hand IMU of the active UE, arriving at a PPV 71% for the remaining forearm IMU. We also used an exhaustive search to automatically identify the most informative number and locations of IMUs on the body. This approach generates classification performances for all combinations of IMUs. This analysis showed that seven IMUs located on the head, trunk, and active UE had the highest PPV (92.5%), confirming the optimal number and configuration identified with domain knowledge.

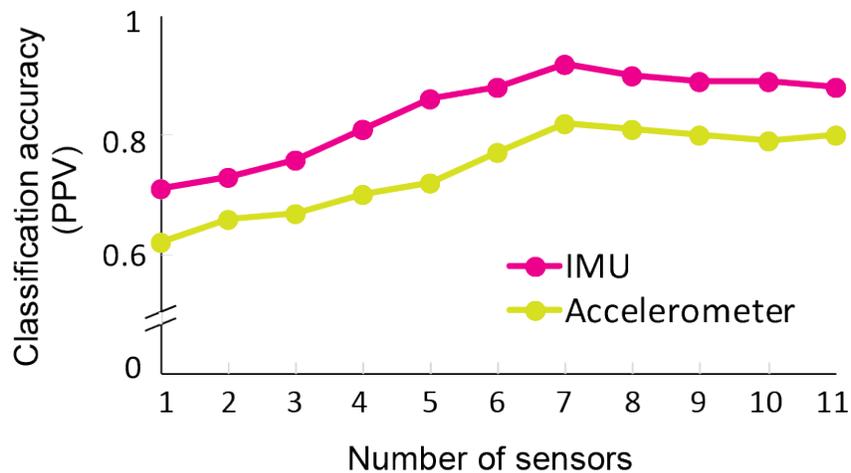

**Figure 5. Classification performance for full and reduced sensor counts.** Performance was computed using LDA and data from with progressively reduced sensor counts. Seven sensors (pelvis, sternum, head, and the active shoulder, upper arm, forearm, and hand) gave the best classification accuracy, with a drop-off at higher and lower sensors counts. IMU data consistently supported higher classification than accelerometer data, achieving PPV 92.5% vs. 82% at 7 sensors.

**Identifying optimal data type.** To finish, we evaluated classification performance using accelerometry data only. As with IMU sensors, seven accelerometers positioned on the head, trunk, and active UE enabled the highest classification performance, with performance drop-offs with more or fewer sensors (removed in the same order as IMUs; Fig. 5). Classification performance using accelerometry data was consistently lower than for IMU data for all sensor configurations (e.g. PPV 84% vs. PPV 92% for seven sensors). Classification performance with accelerometers was lower especially for *reaching* (PPV 77% vs. 93%), which elicited different arm configurations to grasp the objects (e.g. supinating to side-grasp the can versus pronating to overhand grasp the toilet paper roll; Table 4). These findings indicate that IMU data enable a superior level of classification, particularly with more variable motions.

| Primitives | Classification accuracy (PPV) | |
|---|---|---|
| | IMU | Accelerometer |
| Reach | 93% | 77% |
| Transport | 91% | 80% |
| Reposition | 93% | 82% |
| Idle | 92% | 88% |
| Average | 92.5% | 82% |

**Table 4. Primitive-level classification using IMU or accelerometer data.** Classification performance is shown using the 7-sensor configuration (pelvis, sternum, head, and the active shoulder, upper arm, forearm and, hand). Accelerometers had systematically poorer classification performance compared to IMUs across all primitives. Classification performance using accelerometry data was particularly low for reach (PPV 77%) and relatively higher for idle (PPV 88%).

## 5. *Discussion*

Dosing of rehabilitation therapy after stroke remains an elusive clinical challenge. To date, approaches to quantify rehabilitation dose have been limited by impracticality and imprecision. In this study, we aimed to optimize an approach that uses wearable sensors and machine learning algorithms to classify movement primitives, which are summed to quantify dose. We sought to identify—from both performance and pragmatic standpoints—the best machine learning algorithm, sensor configuration, and data type to classify movement primitives in stroke patients. Among the ML algorithms, LDA represented the best balance of classification accuracy and pragmatic implementation. Among sensor configurations, seven sensors on the paretic arm and trunk enabled better classification performance than more or fewer sensors. Among data types, IMU data enabled better classification performance than accelerometers. To our knowledge, this is the first study to compare various ML algorithms, sensor configurations, and data characteristics to automatically classify functional movement primitives in stroke patients.

With recent advancements in wearable sensor technologies, most researchers have focused on activity recognition [30-32]. Only a few that have decomposed complex activity into more fundamental components, but these used vision-based approaches [33, 34]. For example, Sanzari and colleagues identified videotaped movements at single anatomical joints using an unsupervised ML approach [34]. While the study presents a novel approach for identifying human movement at single joints, its relevance to real-world activity could be limited. Human movement typically spans multiple joints in a functional context, creating a highly complex dataset. Furthermore, the approach may not generalize to stroke patients, given the data were generated from healthy controls. In our work, we focus on identifying movement primitives because they represent the fundamental building blocks of activities and provide a finer-grained capture of stroke-impairment movement.

**Optimal performer in classification**. To gauge the real-world applicability of the ML algorithms in classifying movement primitives, we first evaluated their classification performance. Comparing the ML algorithms, we found that LDA and SVM had the highest classification performance, indicted by high PPV (>90%) and AUC (>0.95). These algorithms also had high sensitivities and specificities indicating high true positives and low false positives. LDA shows a high performance because it aims to reduce dimensionality while preserving as much discriminatory information as possible. This approach leads to tight clusters and high separation between the classes. On the other hand, the high performance of SVM arises from the projection of the training data to a high-dimensional space. This approach leads to a maximum separation between classes that may not be possible in the original feature space. Overall, LDA aims to find commonalities within classes of data and difference between classes, whereas SVM aims to find a classification boundary that is farthest from the classes of data. Importantly, these algorithms maximize rigor in the training phase by being less susceptible to noisy or outlier data. LDA accomplishes this by using the clusters centers and not outlying samples to classify, while SVM accomplishes this by using the closest data (i.e., most difficult to discriminate) to define class boundaries. It is worth noting that LDA assumes that the underlying classes are normally distributed (unimodal Gaussians) with the same covariance matrix. If real world movement data are significantly non-Gaussian, the LDA projections may not capture the underlying complex structures required for accurate classification. In this case, classification performance can be improved by allowing the covariance matrices among classes to vary, resulting in a regularized discriminant analysis [35].

By comparison, KNN showed a marginally lower classification performance, possibly due to its susceptibility to noise [36]. In our current setup of KNN, all nearest sample points are given the same weighting. In other words, a noisy sample will be weighed the same as other statistically important samples when assigning a class label. KNN classification performance can be improved with noisy data by choosing an appropriate weighting metric (e.g., inverse squared weighing) [37]. This ensures that samples closer to the test sample contribute more to classifying it. Performance may also be improved by using a variant approach called mutual nearest neighbors, where noisy samples are detected using pseudo-neighbors (neighbors of neighbors) and assigned lower weights [38].

Finally, NBC had the lowest performance compared to other algorithms. This may be attributed to its underlying assumption of conditional independence between data features [39]. This assumption is violated for data streams that are correlated, which negatively influences performance. In our dataset, there is ample correlated data from adjacent sensors on the body, like the hand and wrist. The performance of NBC could be improved by applying principal components analysis (PCA) to the dataset as a pre-processing step, and then training the NBC [40, 41].

Comparing these results with our prior work [14], we found that the four algorithms outperformed the hidden Markov model-logistic regression (HMM-LR) classifier in identifying the movement primitives in stroke patients. Our improved performance may be due in part to differences in training datasets. Our previous study trained the algorithm on healthy controls and tested on stroke patients to examine the generalizability of the model. It is conceivable that if the HMM-LR classifier been trained and tested on stroke patients only, its performance would have been higher.

**Optimal performer in practicality**. Next we sought to determine the most pragmatic algorithm in terms of their computational complexity, i.e., their training and testing times and tuning requirements. Comparing algorithm training times, we found that KNN did not have any computational expense. This is expected, since KNN requires no training and shifts the computational cost to the testing phase. Training times for LDA and NBC grew gradually with dataset size, taking at most minutes. With a small training dataset, LDA outperformed NBC, but required more training time as the dataset increased. This can be explained by the scatter matrix computations and optimization of LDA, which become computationally expensive as the dataset size increases [16]. On the other hand, NBC estimates prior probabilities by counting the number of samples belonging to each class in the training dataset, a process that is computationally faster than matrix computations. By contrast, SVM training time increased quadratically, because finding an optimal hyperplane between classes entails solving a quadratic programming problem [18]. Complex algorithms such as SVM thus require more processing time for large datasets, which limits their use in real-world applications. For example, for a modestly sized study, training times for SVM may be on the order of days. This lag would be prohibitive for rapid tuning, significantly delaying algorithm optimizations.

Comparing algorithm testing times, we found that SVM, LDA, and NBC required less than a millisecond to classify primitives, whereas the testing time for KNN took seconds-minutes and grew linearly with dataset size. This can be explained by how KNN works [42]. During testing, the KNN algorithm searches for the k nearest neighbors around the test sample, i.e., that have similar data characteristics as the test sample. This search is exhaustive and computationally expensive. With increasing samples and dimensionality of the data, the search broadens and takes more time. If an investigator wishes to classify primitives offline, KNN testing times may be acceptable. For applications requiring near- or real-time classification (e.g. for online feedback), the other algorithms should be considered. Alternatively, the classification complexity of KNN can be reduced by selecting an efficient search algorithm (e.g., KD tree) [43], which limits the search space during testing.

Comparing the ease of tuning, we determined that NBC had the lowest parameter complexity and requirement for domain knowledge, whereas SVM had the highest. To address the single tuning parameter of NBC, basic knowledge of statistics is required. KNN has a moderate number of tuning parameters, but they are relatively straightforward to understand and address. LDA has fewer tuning parameters than KNN, but requires moderate domain knowledge to select the amount of regularization allowing the covariance among classes to vary [35]. SVM requires the highest amount of parameter tuning to optimize both classification and practical performance. Building an SVM model requires a deep understanding of statistics, optimization, probability theory, and machine learning [44]. This level of domain knowledge is prohibitive for SVM use in an unsupported research setting.

All told, weighing classification performance and pragmatic implementation, we found that out of the four ML algorithms LDA is the best for primitive detection in IMU data.

**Optimal IMU configuration**. From the hardware side, we sought to identify the optimal sensor location and configuration to facilitate data capture while maintaining high classification accuracy. We showed that seven sensors (not more or fewer) enable optimal classification accuracy, and that the best sensor configuration captures movement only in the moving limb and trunk. This result is expected, given that the participants performed a unimanual task and the sensors on the active arm and trunk captured the movement. Interestingly, accuracy worsened with more sensors, likely because of the increased dimensionality of the dataset. This may cause the ML algorithm to overfit the training data resulting in lower classification accuracy during the testing phase [45]. To maintain the performance while adding more IMUs, more training data will be needed for the ML algorithm to learn an accurate relationship. Finally, we found that if only one sensor were available, the forearm location was the most informative, although classification performance was modest (PPV 71% for an IMU). This location is appealing, given the recent advances in smartwatches that capture movement.

**Optimal data characteristics**. We finally sought to identify movement data characteristics that lead to highest classification performance. We found that accelerometry data consistently generated lower accuracies than IMU data, which is likely due to its fewer dimensions. Although IMUs enable higher classification performance than accelerometers, they also have several practical limitations: a higher risk of electromagnetic drift leading to inaccurate data estimates, a more frequent need for recalibration, a higher consumption of energy [46], and a higher cost. Thus there is a tradeoff between robust capture and practical motion capture. We believe that the benefits of richer data and better classification outweigh the limitations of IMUs. However, if constrained by financial resources or the magnetic noisiness of an environment, accelerometers may be acceptable for coarse UE primitive identification.

**5.1 Limitations and future work**.

Our study has some limitations to be considered. First, our analysis was performed on a small dataset of six mild-to-moderately impaired stroke patients, limiting generalizability to all levels of impairment. To achieve high classification accuracy across the range of stroke impairment, separate ML models may need to be trained for different impairment levels. Second, the activity used in this study was highly structured. The resulting primitives were thus more constrained and consistent than what one might find during real-world performance of ADLs. The algorithms trained on this dataset thus may not generalize to all ADLs. Future work is needed to train and test algorithms on functional primitives with an array of kinematic characteristics.

## 6. Conclusion

In summary, we refined a strategy to precisely and pragmatically quantify movement primitives in stroke patients. We evaluated four off-the-shelf machine learning algorithms, finding that LDA had the best combination of classification performance and pragmatic performance. We also found that seven sensors on the paretic UE and trunk optimized classification, and that IMUs enabled superior classification compared to accelerometers. Future studies may consider implementing our improved approach for classifying movement primitives in stroke patients.

**Declarations:**
- **Ethics approval and consent to participate**: Institutional Review Board-approved testing occurred at Columbia University. Subjects gave written informed consent to participate in this study, in accordance with the Declaration of Helsinki.
- **Consent for publication**: Not applicable
- **Availability of data and material**: The dataset analyzed for the current study are available from the corresponding author on reasonable request.
- **Competing interests**: All authors report no financial or non-financial competing interests.
- **Funding**: The work was supported by K02NS104207 (HS) and K23NS078052 (HS).
- **Authors' contributions**: AP analyzed and interpreted the data and wrote paper. JU collected and coded data. DN created the activities battery and interpreted the data. HS collected, coded data, and interpreted the data and wrote the paper.
- **Acknowledgements**: We would like to acknowledge Dr. Sunil Agarwal for provision of the sensor system and Jorge Guerra for the early machine learning analysis.

## References

[1] B. Ovbiagele *et al.*, "Forecasting the future of stroke in the United States: a policy statement from the American Heart Association and American Stroke Association," *Stroke,* vol. 44, no. 8, pp. 2361-75, Aug 2013.

Parnandi *et al.*                                                                                                           Page **19** of **21**[2]     G. Kwakkel, J. M. Veerbeek, E. E. van Wegen, and S. L. Wolf, "Constraint-induced movement therapy after stroke," *Lancet Neurol,* vol. 14, no. 2, pp. 224-34, Feb 2015.

[3]     D. Mozaffarian *et al.*, "Heart Disease and Stroke Statistics—2016 Update," *A Report From the American Heart Association,* 2015.

[4]     J. Biernaskie, G. Chernenko, and D. Corbett, "Efficacy of rehabilitative experience declines with time after focal ischemic brain injury," *J Neurosci,* vol. 24, no. 5, pp. 1245-54, Feb 4 2004.

[5]     R. J. Nudo, B. M. Wise, F. SiFuentes, and G. W. Milliken, "Neural substrates for the effects of rehabilitative training on motor recovery after ischemic infarct," (in eng), *Science,* vol. 272, no. 5269, pp. 1791-4, Jun 21 1996.

[6]     J. A. Bell, M. L. Wolke, R. C. Ortez, T. A. Jones, and A. L. Kerr, "Training Intensity Affects Motor Rehabilitation Efficacy Following Unilateral Ischemic Insult of the Sensorimotor Cortex in C57BL/6 Mice," *Neurorehabilitation and Neural Repair,* vol. 29, no. 6, pp. 590-598, July 1, 2015 2015.

[7]     C. Bregler, "Learning and recognizing human dynamics in video sequences," in *Computer Vision and Pattern Recognition, 1997. Proceedings., 1997 IEEE Computer Society Conference on*, 1997, pp. 568-574: IEEE.

[8]     T. Flash and B. Hochner, "Motor primitives in vertebrates and invertebrates," *Current Opinion in Neurobiology,* vol. 15, no. 6, pp. 660-666, 12// 2005.

[9]     C. E. Lang, D. F. Edwards, R. L. Birkenmeier, and A. W. Dromerick, "Estimating minimal clinically important differences of upper-extremity measures early after stroke," *Arch Phys Med Rehabil,* vol. 89, no. 9, pp. 1693-700, Sep 2008.

[10]    K. R. Lohse, C. E. Lang, and L. A. Boyd, "Is more better? Using metadata to explore dose-response relationships in stroke rehabilitation," *Stroke,* vol. 45, no. 7, pp. 2053-8, Jul 2014.

[11]    G. Kwakkel *et al.*, "Effects of augmented exercise therapy time after stroke: a meta-analysis," (in eng), *Stroke,* vol. 35, no. 11, pp. 2529-39, Nov 2004.

[12]    C. E. Lang *et al.*, "Observation of amounts of movement practice provided during stroke rehabilitation," (in eng), *Arch Phys Med Rehabil,* vol. 90, no. 10, pp. 1692-8, Oct 2009.

[13]    L. Chen, J. Hoey, C. D. Nugent, D. J. Cook, and Y. Zhiwen, "Sensor-Based Activity Recognition," *Systems, Man, and Cybernetics, Part C: Applications and Reviews, IEEE Transactions on,* vol. 42, no. 6, pp. 790-808, 2012.

[14]    J. Guerra *et al.*, "Capture, learning, and classification of upper extremity movement primitives in healthy controls and stroke patients," (in eng), *IEEE Int Conf Rehabil Robot,* vol. 2017, pp. 547-554, Jul 2017.

[15]    L. Bao and S. Intille, "Activity recognition from user-annotated acceleration data," *Pervasive computing,* pp. 1-17, 2004.

[16]    A. M. Khan, Y.-K. Lee, S. Y. Lee, and T.-S. Kim, "A triaxial accelerometer-based physical-activity recognition via augmented-signal features and a hierarchical recognizer," *IEEE transactions on information technology in biomedicine,* vol. 14, no. 5, pp. 1166-1172, 2010.

[17]    N. Ravi, N. Dandekar, P. Mysore, and M. L. Littman, "Activity recognition from accelerometer data," in *Aaai*, 2005, vol. 5, no. 2005, pp. 1541-1546.

[18]    C. Schuldt, I. Laptev, and B. Caputo, "Recognizing human actions: a local SVM approach," in *Pattern Recognition, 2004. ICPR 2004. Proceedings of the 17th International Conference on*, 2004, vol. 3, pp. 32-36: IEEE.

[19]   Mathworks. (2018). *Analyze and model data using statistics and machine learning*. Available: https://www.mathworks.com/products/statistics.html
[20]   (2018). *scikit-learn Machine Learning in Python*. Available: http://scikit-learn.org/stable/
[21]   A. J. Izenman, "Linear discriminant analysis," in *Modern multivariate statistical techniques*: Springer, 2013, pp. 237-280.
[22]   K. P. Murphy, "Naive bayes classifiers," *University of British Columbia,* vol. 18, 2006.
[23]   M. A. Hearst, S. T. Dumais, E. Osuna, J. Platt, and B. Scholkopf, "Support vector machines," *IEEE Intelligent Systems and their applications,* vol. 13, no. 4, pp. 18-28, 1998.
[24]   J. Fürnkranz, "Round robin classification," *Journal of Machine Learning Research,* vol. 2, no. Mar, pp. 721-747, 2002.
[25]   L. E. Peterson, "K-nearest neighbor," *Scholarpedia,* vol. 4, no. 2, p. 1883, 2009.
[26]   R. Parikh, A. Mathai, S. Parikh, G. C. Sekhar, and R. J. I. j. o. o. Thomas, "Understanding and using sensitivity, specificity and predictive values," vol. 56, no. 1, p. 45, 2008.
[27]   B. JC, "Statistical Analysis and Presentation of Data," in *Evidence-Based Laboratory Medicine; Principles, Practice and Outcomes*, C. Price and R. Christenson, Eds. 2 ed. Washington DC, USA: AACC Press, 2007, pp. 113-40.
[28]   Intel. *Intel® Core™ i7-7600U Processor*. Available: https://ark.intel.com/products/97466/Intel-Core-i7-7600U-Processor-4M-Cache-up-to-3-90-GHz-
[29]   E. W. Weisstein. *Exhaustive Search. From MathWorld--A Wolfram Web Resource.* Available: http://mathworld.wolfram.com/ExhaustiveSearch.html
[30]   R. J. Lemmens, Y. J. Janssen-Potten, A. A. Timmermans, R. J. Smeets, and H. A. Seelen, "Recognizing complex upper extremity activities using body worn sensors," *PLoS One,* vol. 10, no. 3, p. e0118642, 2015.
[31]   D. Biswas *et al.*, "Recognition of elementary arm movements using orientation of a tri-axial accelerometer located near the wrist," *Physiol Meas,* vol. 35, no. 9, pp. 1751-68, Sep 2014.
[32]   N. A. Capela, E. D. Lemaire, and N. Baddour, "Feature selection for wearable smartphone-based human activity recognition with able bodied, elderly, and stroke patients," *PLoS One,* vol. 10, no. 4, p. e0124414, 2015.
[33]   J. Bai, J. Goldsmith, B. Caffo, T. A. Glass, and C. M. Crainiceanu, "Movelets: A dictionary of movement," *Electron J Stat,* vol. 6, pp. 559-578, 2012.
[34]   M. Sanzari, V. Ntouskos, S. Grazioso, F. Puja, and F. Pirri, "Human motion primitive discovery and recognition," *Arxiv,* 2017.
[35]   J. H. Friedman, "Regularized discriminant analysis," *Journal of the American statistical association,* vol. 84, no. 405, pp. 165-175, 1989.
[36]   C. M. Van der Walt and E. Barnard, "Data characteristics that determine classifier performance," 2006.
[37]   D. Bridge, "Classification: k Nearest Neighbours," ed, 2007.
[38]   H. Liu, S. J. J. o. S. Zhang, and Software, "Noisy data elimination using mutual k-nearest neighbor for classification mining," vol. 85, no. 5, pp. 1067-1074, 2012.
[39]   M. Tom, "Generative and discriminative classifiers: Naive bayes and logistic regression," ed, 2005.

[40]   L. Fan and K. L. Poh, "A comparative study of PCA, ICA and class-conditional ICA for naïve bayes classifier," in *International Work-Conference on Artificial Neural Networks*, 2007, pp. 16-22: Springer.

[41]   L. Fan and K. L. Poh, "Improving the naïve Bayes classifier," in *Encyclopedia of artificial intelligence*: IGI Global, 2009, pp. 879-883.

[42]   P. Cunningham and S. J. J. M. C. S. Delany, "k-Nearest neighbour classifiers," vol. 34, no. 8, pp. 1-17, 2007.

[43]   J. L. Bentley, "Multidimensional binary search trees used for associative searching," *Communications of the ACM,* vol. 18, no. 9, pp. 509-517, 1975.

[44]   B. Scholkopf and A. J. Smola, *Learning with kernels: support vector machines, regularization, optimization, and beyond*. MIT press, 2001.

[45]   O. M. Giggins, K. T. Sweeney, and B. Caulfield, "Rehabilitation exercise assessment using inertial sensors: a cross-sectional analytical study," *J Neuroeng Rehabil,* vol. 11, p. 158, Nov 27 2014.

[46]   Q. Liu *et al.*, "Gazelle: Energy-efficient wearable analysis for running," no. 9, pp. 2531-2544, 2017.